\documentclass[letterpaper, 10 pt, conference]{ieeeconf}  

\IEEEoverridecommandlockouts                              

\usepackage{xcolor}  
\definecolor{cvprblue}{rgb}{0.21,0.49,0.74}  
\definecolor{mygreen}{RGB}{0,150,0}
\definecolor{myred}{RGB}{200,0,0}
\definecolor{lightblue}{RGB}{230, 242, 255}

\usepackage[pagebackref,breaklinks,colorlinks,allcolors=cvprblue]{hyperref}
\usepackage{graphicx}
\usepackage{wrapfig}
\usepackage{multirow}
\usepackage{booktabs}
\usepackage[table]{xcolor}
\usepackage{relsize}
\usepackage[most]{tcolorbox}
\usepackage{cuted}
\usepackage{caption}
\usepackage{amsmath} 
\usepackage{amssymb} 
\usepackage{framed}
\usepackage{hyperref}
\usepackage[misc]{ifsym}
\let\labelindent\relax
\usepackage{enumitem}

\title{TacReasoner: A Dynamic Tactile-Language Framework for Interactive Reasoning in Real-World Scenarios}

\author{
Kailin Lyu$^{1,2}$,
Di Wu$^{1}$,
Long Xiao$^{1}$,
Jianning Zeng$^{1}$,
Jianwei He$^{1}$,
Chang Lin$^{1}$,
Lianyu Hu$^{3}$,\\
Lin Shu$^{1,4}$,
Jie Hao$^{1,*}$,
and Ce Hao$^{2,*}$%
\thanks{This work was supported by the Zhongguancun Academy (Grant No.~C20250502) and the Guangzhou Key Research and Development Program (Grant No.~2025B01J4002).}
\thanks{$^{1}$Institute of Automation, Chinese Academy of Sciences, Beijing, China.}
\thanks{$^{2}$Beijing Zhongguancun Academy, Beijing, China.}
\thanks{$^{3}$Nanyang Technological University, Singapore.}
\thanks{$^{4}$Guangdong Institute of Artificial Intelligence and Advanced Computing.}
\thanks{$^*$Jie Hao and Ce Hao are the corresponding authors.}
}

\begin{document}

\maketitle
\thispagestyle{empty}
\pagestyle{empty}

\begin{strip}
\vspace{-2.8cm}
\centering
\includegraphics[width=\textwidth]{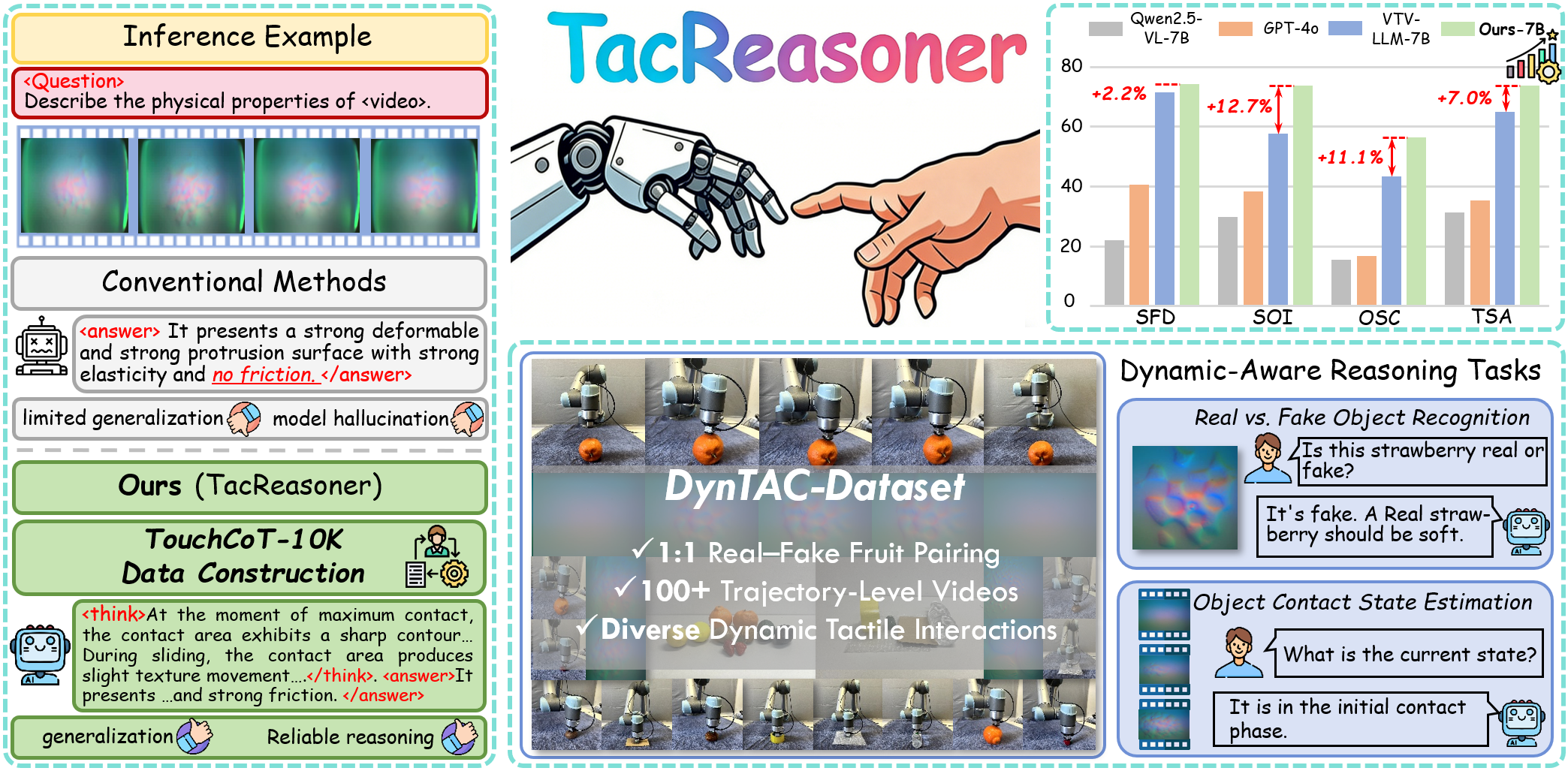}
\captionof{figure}{\textbf{The core contributions of TacReasoner.} (i) We propose a Dynamic-aware Tactile Encoder to strengthen dynamic tactile representation learning. Upon it, we build \textbf{TouchCoT-10K}, a tactile CoT dataset that provides explicit supervision for tactile reasoning. (ii) We further introduce \textbf{DynTac-Bench} to support dynamic perception and tactile interaction tasks. (iii) Across multiple tactile reasoning benchmarks, TacReasoner achieves statistically significant gains over prior methods.}
\vspace{-2mm}
\label{intro}
\end{strip}

\begin{abstract}
Among the five primary human senses, tactile is arguably the most fundamental to survival, as it enables the perception of physical contact and interaction in real-world environments. In this paper, we explore two key challenges of integrating tactile sensing into intelligent systems for multimodal reasoning: (i) insufficient modeling of dynamic tactile signals, which restricts reasoning over temporally evolving properties, and (ii) hallucination in tactile foundation models caused by the absence of explicit reasoning mechanisms, leading to unstable real-world inference. To address these challenges, we propose TacReasoner, a dynamic tactile-language framework for interactive reasoning in real-world scenarios. First, TacReasoner incorporates a Dynamic-aware Tactile Encoder to enhance the perception and representation of dynamic tactile signals. More importantly, we introduce TouchCoT-10k, the first tactile chain-of-thought dataset for structured reasoning over tactile inputs. Upon it, we establish DynTac-Bench to systematically evaluate dynamic tactile perception and real-world commonsense reasoning. Experimental results demonstrate that TacReasoner achieves competitive performance against state-of-the-art models across multiple datasets. Notably, despite using only 7B parameters, TacReasoner outperforms the 14B VTV-LLM model on most subtasks, highlighting its effectiveness and efficiency in tactile commonsense reasoning.
\end{abstract}

\section{INTRODUCTION}
\label{intro}
For humans, tactile is a crucial sense that provides physical information beyond what vision can provide (e.g., material properties, texture information), especially during occlusion \cite{tactile1,tactile2,human11}. This in turn improves our ability to perform physical reasoning and act in our world. For example, when a person touches a hard grain of rice, they can infer that it is under-ripe; similarly, when touching a soft sponge, they can deduce that it is suitable for wiping based on the context. Therefore, integrating tactile information into the commonsense reasoning framework and developing tactile reasoning capabilities for real-world scenarios is crucial for advancing embodied intelligence and holds significant practical importance.

In recent years, several studies have explored integrating tactile perception with large language models to enable robots to perform tactile understanding tasks guided by natural language prompts. For example, Octopi fine tunes a CLIP encoder \cite{clip} with predefined attributes to learn tactile representations and adapts Vicuna for downstream reasoning \cite{octopi}. Octopi 1.5 further incorporates retrieval augmented generation to improve attribute retrieval accuracy \cite{octopi15}. VTV-LLM leverages VideoMAE to learn tactile video representations \cite{vtv}. Despite these advances, \textbf{existing approaches still struggle to achieve reliable and robust tactile reasoning in real world environments}, leaving the final step to empower tactile perception for real-world robotic applications. Their core limitations can be summarized into two key issues:

\noindent \textbf{\textit{(i)}} First, regarding tactile representation, existing methods generally lack explicit modeling of dynamic temporal dependencies in tactile signals and cannot capture physical contact process \cite{octopi,octopi15}. Although some works treat tactile signals as video sequences for reconstruction \cite{vtv}, their objectives are limited to pixel-level recovery and static attribute classification in key contact regions. The supervision signals focus on appearance fidelity rather than physical variations. Such methods fail to characterize cross-frame deformation propagation, shear accumulation, and slip variations during contact, preventing the model from learning genuine dynamic physical laws. From the perspective of neurobiological mechanisms \cite{th2}, human tactile perception relies on the integration of continuous temporal signals by dynamic mechanoreceptors. Existing methods violate this core perceptual principle and thus cannot achieve reliable physical reasoning.

\noindent \textbf{\textit{(ii)}} In addition, regarding tactile reasoning, learning paradigms that rely on static attributes and fixed question-answering templates induce semantic hallucination and limit generalization ability. Current datasets are mostly constructed around limited predefined attributes and trained with template-based queries, such as the recently proposed PHYSICLEAR \cite{octopi} and VTV-150K \cite{vtv}. This forces models to learn shallow and spurious semantic correlations instead of causal reasoning. Meanwhile, tactile sensors lack unified standards \cite{unitouch}, leading to large discrepancies in imaging mechanisms and physical responses. Without explicit constraints on intermediate reasoning steps, models easily overfit to sensor-specific superficial patterns rather than general physical laws. This further causes severe semantic hallucination when facing novel objects, heterogeneous sensors, and unstructured environments, failing to meet the requirements of robotic applications in real‑world scenarios.

To address these challenges, we propose a novel tactile-language framework, \textbf{TacReasoner}, as shown in Fig.~\ref{intro}. First, inspired by biological neuroscience, we model the dynamic neural mechanisms underlying human tactile perception and incorporate a dynamic perception module to enhance tactile sensing capability. Second, we construct the first chain-of-thought dataset for tactile reasoning and optimize the reasoning process to move beyond reliance on discrete attribute labels. This design enables more efficient reasoning in real world scenarios and effectively mitigates semantic hallucination in tactile reasoning. We conduct comprehensive evaluations on tactile attribute inference, comparative tasks, and the more challenging tactile reasoning tasks introduced in this work. Extensive experimental results demonstrate that TacReasoner unifies tactile perception and reasoning through the joint optimization of data and architecture, achieving superior reasoning accuracy over state-of-the-art methods and facilitating reliable and robust reasoning in real-world scenarios. In this work, our main contributions are as follows:

\begin{itemize}[leftmargin=*, itemsep=0pt, parsep=0pt, topsep=2pt, partopsep=0pt]
  \item \textbf{Framework}: We propose \textbf{TacReasoner}, a framework that models dynamic perceptual mechanisms to approximate human tactile perception and enhance the capability of large language models to understand the tactile modality.
  \item \textbf{Datasets}: We construct the first chain of thought dataset for tactile attribute reasoning---\textbf{TouchCoT-10k}, and introduce a suite of challenging benchmark tasks for tactile reasoning, named \textbf{DynTAC-Bench}. These benchmarks facilitate improved tactile reasoning and provide a rigorous evaluation of tactile-language models.
  \item \textbf{Practice}: TacReasoner outperforms existing tactile-language models across multiple mainstream datasets and subtasks, and demonstrates more reliable physical reasoning, making it more suitable for robotic interaction in real-world environments. We hope this work encourages further discussion and exploration in the field and provides new insights for future research on tactile intelligence.
\end{itemize}

\section{Related Work}

\subsection{Visuo-tactile Perception}
\label{Visuo-tactile Perception}
In recent years, tactile sensing has evolved from early sensors that measured only basic physical properties to sophisticated vision-based systems capable of capturing high resolution contact information \cite{tacsensor2,tacsensor3}. Visuo-tactile sensors have attracted significant attention due to their ability to record high resolution spatiotemporal deformations of contact surfaces. Building on these, many studies leverage visuo-tactile sensing to infer multidimensional tactile properties, enabling dexterous manipulation tasks such as material classification \cite{taob,touchformer}, grasping \cite{tacgrasp,grasp2}, and insertion \cite{tacin,tacman2}. Recent research has shifted toward representation learning for tactile data. Existing approaches leverage vision inspired representation learning, employ visual self-supervised objectives for fine grained feature extraction \cite{unitouch,t3}, utilize tactile video reconstruction for feature learning and achieve semantic level understanding through multimodal alignment with vision and language \cite{vtv}. To address sensor heterogeneity, some studies adopt joint training and alignment strategies to enforce cross sensor representation consistency \cite{unitouch,anytouch}. In contrast to prior work, we focus on modeling the dynamic characteristics of tactile signals and unlocking the potential of tactile representations for complex tactile reasoning.

\subsection{Tactile-Language Model}
\label{Tactile-Language Model}
Multimodal large language models jointly model language and visual information, substantially enhancing cross modal reasoning and advancing research paradigms \cite{mllm}. Early studies primarily focused on vision  language models \cite{qwen,llava}. More recently, researchers have begun leveraging the reasoning and comprehension capabilities of large language models to model tactile signals \cite{octopi,octopi15,vtv,touchthinker}, gradually establishing a tactile-language modeling paradigm for embodied interaction. Yu et al \cite{octopi}. proposed the Octopi framework and constructed the PHYSICLEAR dataset to support basic tactile commonsense reasoning. Building on this foundation, Octopi 1.5 \cite{octopi15} introduces a retrieval augmented generation module that retrieves similar objects from a database to improve prediction performance. VTV-LLM \cite{vtv} focuses on visuo-tactile video modeling and collects cross sensor data to expand task coverage. Although prior work has achieved notable progress, it does not fully account for the dynamic interactive nature of tactile signals or the issue of semantic hallucination during reasoning. In this work, we systematically model the dynamic interaction characteristics of tactile perception and construct the first chain of thought dataset for tactile reasoning to supervise intermediate reasoning processes. This design effectively mitigates hallucination and enhances tactile reasoning in real world settings.

\section{Method}
\label{method}

\begin{figure*}[!t] 
    \centering 
    \includegraphics[width=\textwidth]{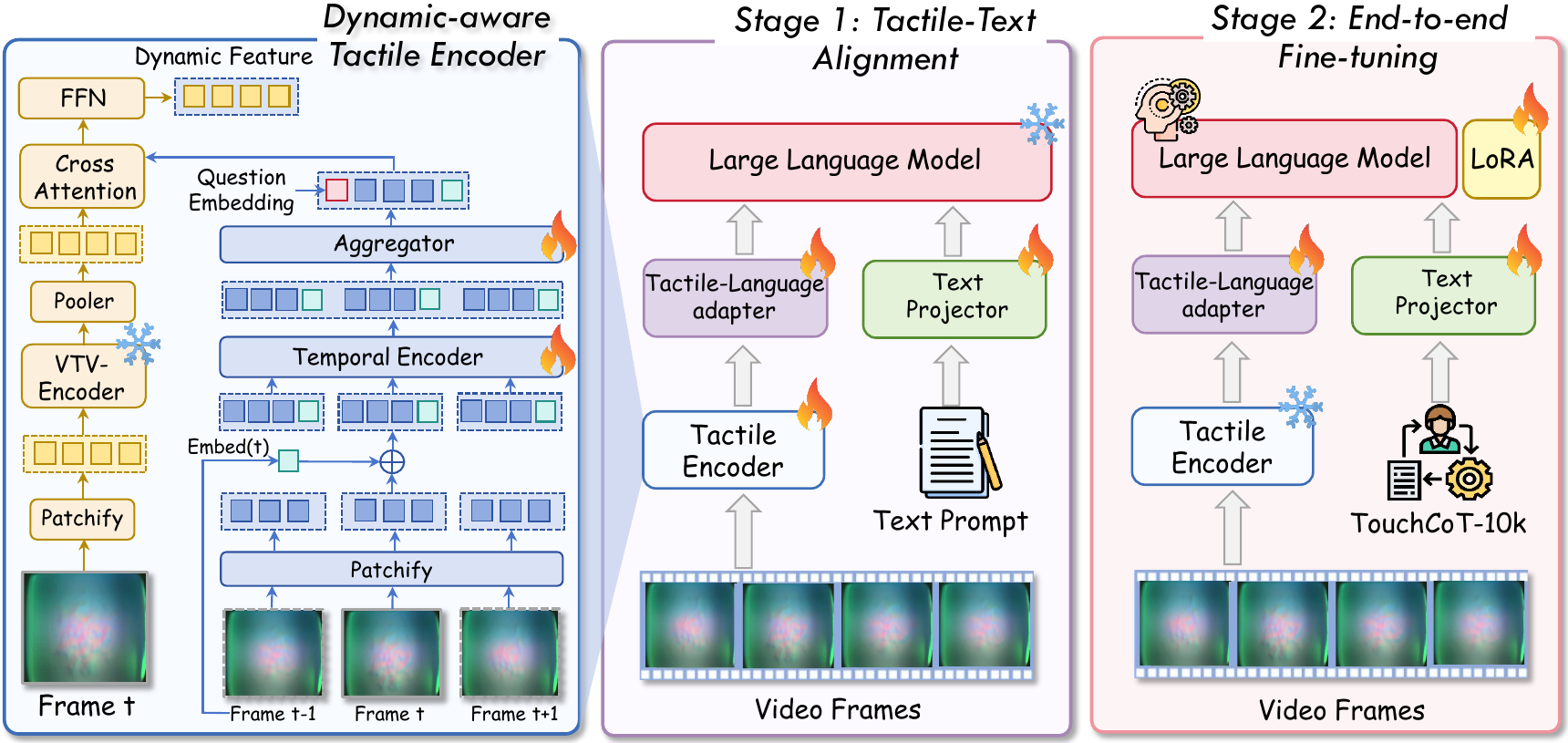}    \caption{\textbf{The framework of TacReasoner.} TacReasoner consists of a tactile encoder, a tactile language adapter, and a large language model. The training procedure follows a two stage strategy. \textit{In Stage I}, we train the tactile encoder and the tactile-language adapter to adapt the large language model to tactile inputs. \textit{In Stage II}, we conduct end to end fine tuning on TouchCoT 10k to activate the reasoning capability of the model and generate accurate tactile descriptions.} 
    \label{pipline}
    \vspace{-3mm}
\end{figure*}

As shown in Figure~\ref{pipline}, TacReasoner integrates temporal and state information into the tactile-language reasoning framework, enabling dynamic perception and structured modeling of tactile interactions. By explicitly modeling the temporal evolution and causal dependencies inherent in the contact process, TacReasoner progressively refines intermediate reasoning representations, thereby enhancing tactile understanding. The following sections present the data generation pipeline, the dynamic-aware tactile encoder, and the overall training paradigm.

\subsection{Data Generation}
\label{Data Generation}

\begin{figure}[!t] 
    \centering 
    \includegraphics[width=\columnwidth]{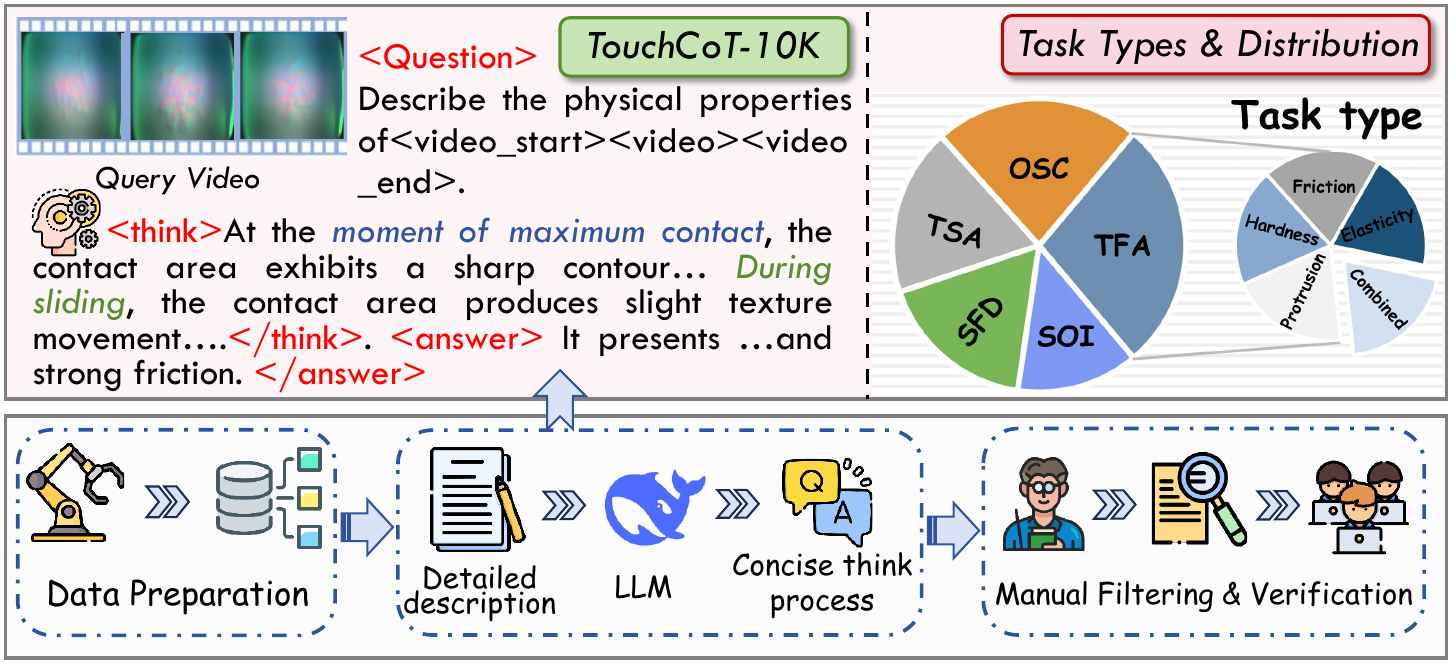} 
    \caption{\textbf{Overview of the TouchCoT-10K construction pipeline and task taxonomy}.}
    \label{method1}
    \vspace{-4mm}
\end{figure}

The field of tactile-language reasoning suffers from a lack of high quality prompt tuning datasets, which fundamentally constrains the ability of models to perform reliable commonsense reasoning and decision making in real world scenarios. Prior work primarily relies on attribute classification or template-based question-answering, providing only discrete labels or brief responses, and lacks structured modeling of contact dynamics and causal dependencies, resulting in shallow semantic associations. Furthermore, as discussed in Section~\ref{intro}, tactile sensors are not standardized, and significant differences in color, size, and artifacts exist between tactile images captured by different sensors \cite{tacsensor3}. This disparity leads models to potentially rely on fixed optical and visual features for reasoning. Due to these limitations, models are prone to increased semantic hallucination and constrained generalization when faced with diverse objects and heterogeneous tactile signal distributions. To address this gap, we have constructed \textbf{TouchCoT-10k}, a high-quality tactile reasoning chain-of-thought dataset based on the intrinsic state changes of objects during tactile dynamic interactions. The detailed pipeline generation process is provided in Figure~\ref{method1}.



\textbf{Step 1: Data Preparation.} We collect and organize tactile video samples acquired through a standardized exploration procedure, ensuring that each sequence includes the indentation to maximum contact and sliding interaction phases. Task metadata are systematically structured, including question type labels and step indicators, to distinguish reasoning tasks and enforce consistent, learnable output formats.

\textbf{Step 2: Chain-of-Thought Generation.} In this stage, tactile perception tasks are formalized as cross modal question answering and descriptive generation. We employ template based prompts to guide a large language model (e.g., deepseek \cite{deepseek}), to perform stepwise reasoning grounded in deformation evolution, contact area variation, and geometric cues derived from tactile videos. Outputs are standardized into structured {\small \texttt{<think></think>}} and {\small \texttt{<answer></answer>}} formats. The reasoning process is aligned with the standardized interaction phases, leveraging interpretable cues from maximum contact and sliding stages to infer physical properties.

\textbf{Step 3: Manual Filtering.} After generation, we conduct rigorous consistency checks and manual auditing to ensure data quality. Samples are filtered if the reasoning chain fails to cover key interaction stages or if there are semantic inconsistencies between the reasoning process and the final answer. This procedure ensures logical coherence between reasoning chains and final attribute prediction.

\textbf{Step 4: Formatting and Data Integration.} In the final stage, we organize question type annotations, tactile video inputs, structured prompts, reasoning chains, and final answers into a unified format to construct training ready chain of thought samples. Each instance follows an explicit output structure defined as
CoT = {\small \texttt{...<think>...</think><answer> ... </answer>}}.
This standardized integration results in the high quality TouchCoT-10k dataset, which supervises TacReasoner to learn dynamic tactile reasoning and enhances tactile understanding and decision reliability in real-world scenarios.



\subsection{Dynamic-aware Tactile Encoder}
\label{Dynamic-aware Tactile Encoder}

As discussed in Section~\ref{intro}, tactile representation fundamentally arises from dynamic interaction processes. Although existing tactile-language models can parse multi-frame tactile sequences and generate outputs \cite{octopi,vtv}, they lack explicit modeling of contact temporal evolution, resulting in insufficient capture of dynamic features. Addressing these limitations is essential for the reliable deployment of tactile language models in real-world scenarios.

To address the aforementioned issues, we introduce the Dynamic-aware Tactile Encoder to explicitly model temporal dynamics in tactile interactions while preserving stable appearance semantics. The module integrates spatial morphology and contact dynamics without increasing the input complexity of the language model. As shown in Figure~\ref{pipline}, we adopt a decoupled design that separately models appearance and temporal branches, followed by attention-based feature fusion for unified enhancement. Given a tactile video sequence $ V=\{I_t\}_{t=0}^{T} $, where 
$ I_t \in \mathbb{R}^{H\times W\times 3} $, 
we first extract appearance features using a pretrained VTV encoder \cite{vtv}:
\begin{equation}
F_{\text{app}} = f_{\text{enc}}(V)
= \text{ViT}\left(\left\{\text{Patch}(I_t) + \text{TE}(t)\right\}_{t=0}^{T}\right),
\end{equation}
where $\text{Patch}(\cdot)$ denotes patch embedding and $\text{TE}(t)$ represents temporal positional encoding. 
The global CLS token is adopted as the appearance representation 
$ F_{\text{app}} \in \mathbb{R}^{B\times C} $, 
and the encoder is frozen during training to preserve stable geometric semantics.

To model contact dynamics, we introduce a trainable temporal branch. 
Given raw frames $ X \in \mathbb{R}^{B \times T \times 3 \times H \times W} $, 
we compute frame differences $ \Delta I_t = I_t - I_{t-1} $ 
to explicitly capture inter-frame deformation. 
The difference sequence is encoded by a lightweight temporal encoder:
\begin{equation}
F_{\text{temp}} = \mathrm{Enc}_{\text{temp}}\left(\{\Delta I_t\}_{t=1}^{T}\right),
\quad
F_{\text{temp}} \in \mathbb{R}^{B \times L \times C},
\end{equation}
Notably, to incorporate task-specific semantic information, we encode the question text into embeddings and inject them as conditional signals to enable question-guided temporal attention. The temporal features are then aggregated:
\begin{equation}
\bar{F}_{\text{temp}} = \mathrm{Aggregator}(F_{\text{temp}}).
\end{equation}

Then, we fuse appearance and temporal representations via cross-attention:
\begin{equation}
F_{\text{attn}} = \mathrm{Attention}(Q = F_{\text{app}}, K = \bar{F}_{\text{temp}}, V = \bar{F}_{\text{temp}}),
\end{equation}
\begin{equation}
F_{\text{enh}} = \mathrm{FFN}(F_{\text{attn}}) + F_{\text{app}}.
\end{equation}

Finally, the enhanced tactile representation is projected into the language model embedding space:
\begin{equation}
E_V = W_2 \, \mathrm{GELU}(W_1 F_{\text{enh}} + b_1) + b_2.
\end{equation}

\subsection{Overall training paradigm}
\label{Overall training paradigm}
In the previous section, we introduced the dynamic-aware tactile encoder, which achieves efficient tactile representation through dynamic perception. We next present a two-stage training paradigm that allows the LLM to integrate tactile and linguistic embeddings for multimodal reasoning and response generation.

\begin{figure*}[!t] 
    \centering 
    \includegraphics[width=\textwidth]{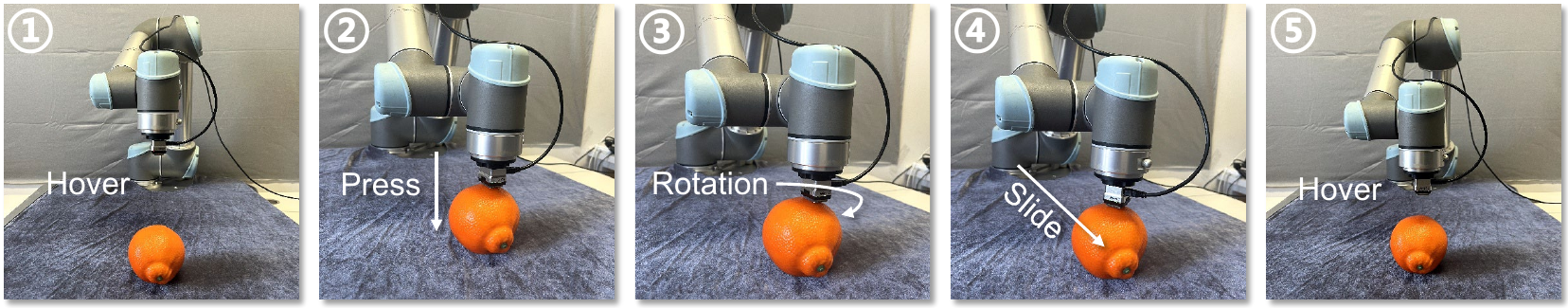}    \caption{\textbf{DynTAC data acquisition process.} The robotic arm initializes away from the object surface, performs pressing, rotation, and sliding upon contact, and retracts, generating a tactile video of approximately five seconds. It then repositions and repeats the procedure to obtain five videos in total. All remaining data are collected under the same protocol.} 
    \label{dataset2}
    \vspace{-3mm}
\end{figure*}

\textbf{Stage I: Tactile-Text Alignment.} At this stage, the objective is to adapt tactile tokens produced by the tactile encoder to the LLM, enabling effective interpretation of tactile inputs. A tactile-language adapter maps tactile tokens into the LLM text embedding space by treating tactile patches as pseudo-text tokens. The adapter is trained using paired tactile images and language descriptions from the raw instruction data of VTV-150K \cite{vtv}, while the LLM remains frozen \cite{qwen}.

The model is optimized by minimizing the cross-entropy loss over the generated token sequence $Y = [y_1, y_2, \dots, y_M] \in \mathbb{R}^{M \times D}$:
\begin{equation}
\mathcal{L}_{\mathrm{ce}} =
- \mathbb{E}_{(Y_i \mid V, T_{<i}) \sim \pi_\theta}
\left[
\log \pi_\theta(Y_i \mid V, T_{<i})
\right],
\end{equation}
where teacher forcing is employed during training, and $\pi_\theta(Y_i \mid V, T_{<i})$ denotes the probability of predicting token $Y_i$ conditioned on tactile representation $V$ and the preceding $i-1$ target tokens $T_{<i}$.

\textbf{Stage II: End-to-end Supervised Fine-Tuning.} In this stage, we activate the model’s tactile reasoning capability through LoRA-based supervised fine-tuning \cite{lora} on the TouchCoT-10K dataset. The objective is to enable the model to learn structured tactile reasoning patterns while maintaining logical consistency between intermediate chain-of-thought representations and final attribute predictions. To preserve stable perceptual representations, we freeze the tactile encoder, and fine-tune the tactile-language adapter and the LLM using instruction data. Parameter-efficient adaptation is applied to the self-attention layers via LoRA. Training samples follow a triplet format $(V, p, O)$, where $V$ denotes the tactile video input, $p$ the textual prompt, and $O$ the structured output sequence. Each output adopts a standardized chain-of-thought format:

$O={\small \texttt{<think> T </think><answer> A </answer>}}$,

\noindent where $T$ represents the reasoning process and $A$ the final attribute prediction. Through high-quality CoT supervision, the model learns to align dynamic tactile evidence with structured reasoning outputs. The optimization objective minimizes the conditional cross-entropy loss over the generated token sequence $Y=[y_1,\dots,y_M]\in\mathbb{R}^{M\times D}$:
\begin{equation}
\mathcal{L}_{\mathrm{SFT}}
=
-\mathbb{E}_{(V,p,O)\sim\mathcal{D}_{\mathrm{TouchCoT}}}
\sum_{i=1}^{M}
\log \pi_{\theta}(y_i \mid E_V, p, y_{<i}),
\end{equation}
where $E_V$ denotes the enhanced tactile embedding produced by the dynamic-aware tactile encoder, and teacher forcing is applied during training. This stage establishes tactile-language alignment for structured reasoning.

\section{DynTAC-Bench}
\label{DynTAC-Bench}

\begin{figure}[!t] 
    \centering 
    \includegraphics[width=\columnwidth]{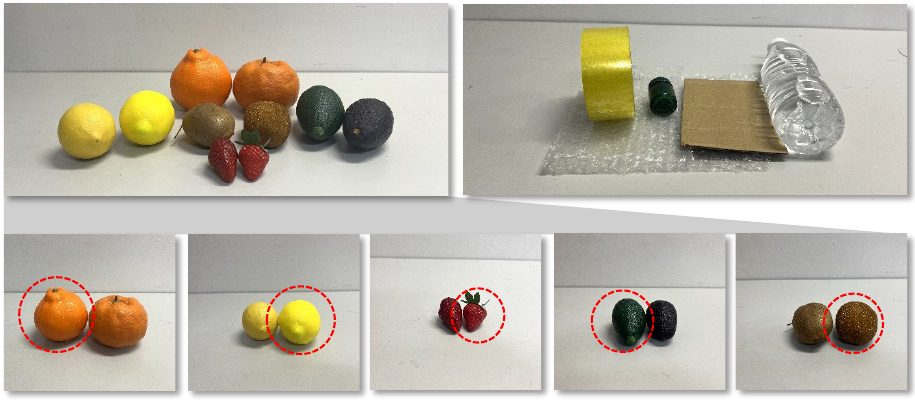} 
    \caption{\textbf{Overview of the DynTAC Dataset.} (1) The top left panel shows five 1:1 real and fake fruit pairs. The bottom row presents detailed views, with red dashed circles marking fake objects that are indistinguishable from real ones under visual inspection or static touch. (2) The top right panel displays five categories of daily objects.} 
    \label{dataset1}
    \vspace{-4mm}
\end{figure}

To evaluate model reasoning in dynamic tasks and real-world scenarios, We propose a new benchmark, \textbf{DynTAC-Bench}, as shown in Fig.~\ref{dataset1}, which covers tasks ranging from fundamental tactile perception to complex reasoning. We introduce its task taxonomy and dataset construction below.

\subsection{Dataset Construction}
\label{Dataset Construction}

We construct the high-quality chain-of-thought dataset TouchCoT-10K based on the raw VTV-150K dataset, as detailed in Section~\ref{Data Generation}. To further support dynamic perceptual reasoning, we introduce the DynTAC dataset (Figure~\ref{dataset1}), which comprises five 1:1 real and fake fruit pairs and five categories of daily objects. The fake fruits are highly realistic and indistinguishable from real ones through visual inspection or static touch, necessitating dynamic tactile perception for reliable discrimination. The daily object categories exhibit substantial variation in material properties and appearance, enabling comprehensive evaluation of model generalization in real-world tactile perception. Tactile data are collected using a UR5 robotic arm under a standardized interaction protocol, as shown in Figure~\ref{dataset2}. The data acquisition process incorporates diverse dynamic tactile interaction modalities, including pressing, rotation, and sliding, thereby generating structured temporal signals that support the evaluation of dynamic perception and reasoning tasks.

\begin{table*}[!t]
\centering
\caption{Performance comparison of TacReasoner against seven state-of-the-art methods on the VTV-150K \cite{vtv} dataset. The evaluation covers different tasks, with results reported in percentages (\%) and the boldface indicates the best performance. Notably, TacReasoner-7B outperforms the 14B-parameter VTV-LLM on most sub-tasks.}

\renewcommand{\arraystretch}{1.15}
\resizebox{\textwidth}{!}{
\begin{tabular}{l|ccccc|cccc|c}
\hline
Models & Hardness & Protrusion & Elasticity & Friction & Combined & SFD & SOI & OSC & TSA & Average \\
\hline
GPT-4o \cite{gpt4o} & 34.7 & 32.6 & 32.6 & 18.7 & 2.1 & 40.9 & 38.4 & 16.6 & 36.0 & 28.0 \\
Gemini-2.5-Pro-Exp \cite{gemini}& 36.2 & 34.7 & 39.1 & 21.0 & 4.3 & 42.6 & 29.4 & 18.5 & 40.0 & 29.5 \\
LLaVA-OneVision-7B \cite{llava}& 27.5 & 32.6 & 26.0 & 20.2 & 0.7 & 40.9 & 28.2 & 11.7 & 30.0 & 24.2 \\
LLaVA-Video-Qwen2-7B \cite{llava-video}& 30.4 & 29.7 & 28.9 & 18.1 & 2.1 & 33.6 & 29.4 & 17.2 & 36.0 & 25.0 \\
InternVL2.5-VL-8B \cite{internvl}& 18.1 & 23.9 & 21.0 & 13.7 & 0.0 & 24.5 & 17.9 & 11.1 & 24.0 & 17.1 \\
VideoLLaMA3-7B \cite{videollama} & 15.2 & 21.7 & 14.4 & 10.8 & 0.0 & 11.4 & 12.8 & 7.4 & 20.0 & 12.6 \\
Qwen2.5-VL-7B \cite{qwen}& 25.3 & 28.9 & 17.3 & 15.9 & 1.4 & 22.9 & 28.2 & 16.0 & 30.0 & 20.6 \\
\hline
VTV-LLM-7B & 73.9 & 75.0 & 67.3 & 56.5 & 35.6 & 71.3 & 57.6 & 43.2 & 64.0 & 60.4 \\
VTV-LLM-14B & 72.1 & 78.2 & 68.1 & 52.8 & 38.2 & 72.1 & 59.7 & 45.9 & 72.0 & 62.1 \\
\hline
\rowcolor{lightblue} \textbf{TacReasoner-7B} & \textbf{78.16} & \textbf{77.82} & \textbf{76.54} & \textbf{58.46} & \textbf{40.35} & \textbf{73.56} & \textbf{70.32} & \textbf{54.27} & \textbf{71.0} & \textbf{66.7} \\
$\Delta$ (with VTV-LLM-7B) & \color{mygreen}+4.26 & \color{mygreen}+2.82 & \color{mygreen}+9.24 & \color{mygreen}+1.96 & \color{mygreen}+4.75&  \color{mygreen}+2.26 & \color{mygreen}+12.72 & \color{mygreen}+11.07 & \color{mygreen}+7.0 & \color{mygreen}+6.3   \\
\textbf{TacReasoner-14B} & 77.84 & 78.62 & 79.76 & 54.75 & 41.58 & 76.93 & 74.89 & 58.41 & 75.0 & 68.6 \\
\hline
\end{tabular}
}
\vspace{-2mm}
\label{tab1}
\vspace{-2mm}
\end{table*}

\subsection{Task Category}
\label{Task Category}
\textbf{Fundamental Property Understanding.} It requires recognizing and describing an object’s basic physical properties, including hardness, roughness, and texture. The model must perceive these attributes from tactile signals and convert them into human-interpretable textual outputs.

\textbf{Commonsense-Driven Reasoning}. This subtask extends beyond tactile perception by requiring the integration of external commonsense knowledge for reasoning. It involves understanding an object’s behavior, function, or usage in specific contexts and making high-level decisions grounded in tactile evidence. Specifically, it encompasses Surface Feature Distinction (SFD), Surface Optimality Identification (SOI), Object Sensation Correlation (OSC), and Tactile Scenario Analysis (TSA). SFD compares tactile properties across objects to determine relative differences. SOI analyzes multiple surfaces to identify which exhibits the highest degree of a specific attribute. OSC associates tactile perceptual cues with the identity of a real-world object. TSA evaluates the application of haptic knowledge to real-world scenarios that require physical reasoning. Notably, the TSA task is excluded from the training set.

\textbf{Dynamic-Aware Reasoning.} This subtask focuses on modeling the temporal dynamics of tactile interactions, requiring the model to perform state estimation and semantic reasoning based on sequential contact information. In \textbf{\textit{Real vs. Fake Object Recognition}}, the model infers whether dynamically perceived tactile attributes are consistent with the physical characteristics of authentic objects. In \textbf{\textit{Object Contact State Estimation}}, the model analyzes continuous tactile variations to determine the current stage of interaction.

\section{Experiments}
\label{Experiments}

In this section, we evaluate the proposed method in terms of physical property prediction and tactile reasoning. We design a series of experiments to address the following questions: \textbf{(1)} Can TacReasoner improve physical property prediction and support reasoning in everyday scenarios? \textbf{(2)} Does TacReasoner enhance dynamic perception and achieve more accurate performance on dynamic-aware reasoning tasks? \textbf{(3)} Can TacReasoner generalize its understanding of physical properties to unseen daily objects?

\begin{figure}[!t] 
    \centering 
    \includegraphics[width=\columnwidth]{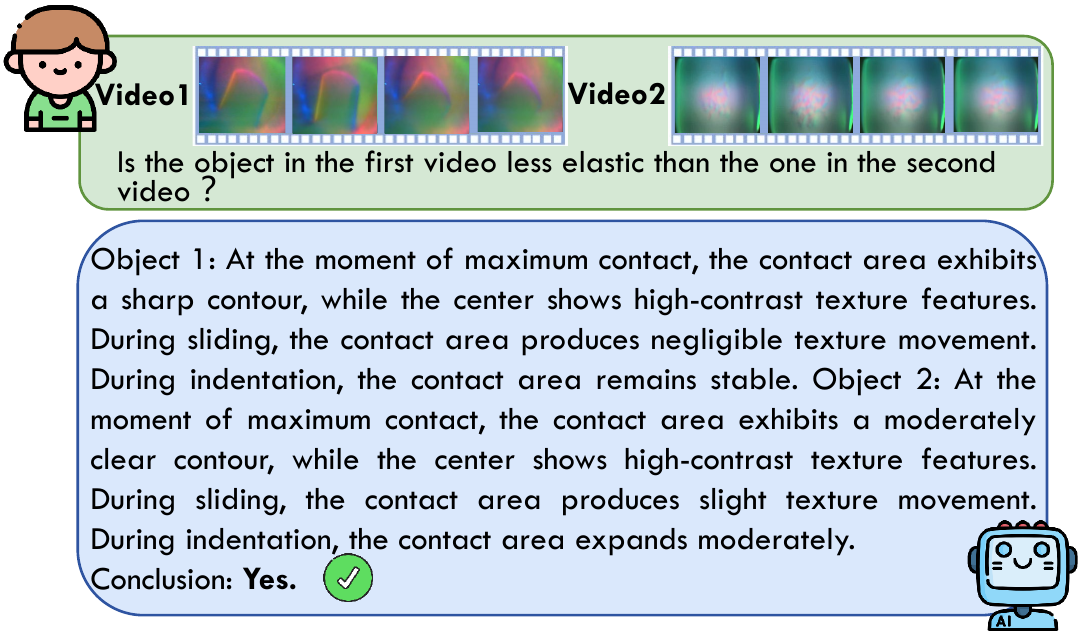} 
    \caption{\textbf{The task examples and the corresponding predictions from TacReasoner.} Video1 shows a wrench, and Video2 shows an avocado. TacReasoner accurately answers questions across diverse sensor inputs by reasoning over intrinsic tactile physical states.} 
    \vspace{-4mm}
    \label{exampe1} 
\end{figure}

\subsection{Implementation Details}
\label{Implementation Details}
All models are trained on two NVIDIA A100-80G GPUs. The training procedure follows our proposed two-stage paradigm. In Stage I, the Tactile-Language Adapter are trained using the raw instruction data from VTV-150K \cite{vtv} with the AdamW \cite{adamw} optimizer and a learning rate of $2 \times 10^{-4}$. In Stage II, the adapter module and the LoRA parameters of the LLM are fine-tuned using TouchCoT-10K dataset with the AdamW optimizer. The learning rates for both the adapter module and the LLM are set to $2 \times 10^{-4}$. For LoRA, we adopt a scaling factor of 256, a rank of 128, and a maximum of 10{,}000 training steps. To evaluate model performance, we construct an independent test set comprising 545 question-answer pairs for \textbf{novel objects not seen during training}. The LLM backbone is based on Qwen 2.5 \cite{qwen}, and we experiment with two model scales containing 7B and 14B parameters.

\subsection{Tactile-grounded physical property prediction}
\label{Tactile-grounded physical property prediction}
We quantitatively compare TacReasoner against VTV-LLM~\cite{vtv}, two proprietary multimodal models (GPT-4o~\cite{gpt4o}, Gemini-2.5-Pro-Exp~\cite{gemini}), and five open-source video-based VLMs: LLaVA-OneVision-7B~\cite{llava}, LLaVA-Video~\cite{llava-video}, InternVL2.5-VL-8B~\cite{internvl}, VideoLLaMA3-7B~\cite{videollama}, and Qwen2.5-VL-7B~\cite{qwen}. All evaluations are performed on 500 question answer pairs sampled from the VTV-150K dataset \cite{vtv}. Since most of the VLM models have parameters around 7B, we adopt TacReasoner-7B to ensure a fair comparison. To guarantee result robustness, we report the average performance over three runs with different random seeds. The experimental results are summarized in Table~\ref{tab1}. The results demonstrate that on the Tactile Feature Analysis (TFA) task, TacReasoner-7B outperforms VTV-LLM-7B by \textbf{4.26\%}, \textbf{2.82\%}, \textbf{9.24\%}, and \textbf{1.96\%} in hardness, bumpiness, elasticity, and friction prediction, respectively, validating the effectiveness of our approach. Notably, elasticity prediction improves by \textbf{nearly 10\%}, and friction also shows measurable gains. As both elasticity and friction are dynamic properties, these improvements highlight that TacReasoner effectively captures dynamic tactile features and leverages them to enhance tactile understanding.

\subsection{Tactile scenario reasoning}
\label{Tactile scenario reasoning}
To further evaluate the reasoning capability of TacReasoner, we assess its performance on SFD, SOI, OSC, and TSA tasks, which require the model to leverage physical properties and perform comparison, matching, and inference. As shown in Table~\ref{tab1}, TacReasoner achieves larger gains on these reasoning tasks than on TFA task. Specifically, compared with VTV-LLM-7B, TacReasoner improves performance by \textbf{2.26\%}, \textbf{12.72\%}, \textbf{11.07\%} and \textbf{7.0\%} on SFD, SOI, OSC, and TSA, respectively, demonstrating strong competitive capability. In addition, we provide a qualitative SFD example in Figure~\ref{exampe1} to illustrate TacReasoner effectively performing commonsense physical reasoning. Remarkably, TacReasoner-7B surpasses VTV-LLM-14B \cite{vtv} in overall performance and on most subtasks, despite using significantly fewer parameters. This improvement stems from its effective tactile representation learning and its ability to elicit genuine reasoning in tactile-language models.

\begin{table}[!t]
\centering
\renewcommand{\arraystretch}{1.1}
\setlength{\tabcolsep}{3pt}
\caption{\textbf{Results on the tactile dynamic-aware reasoning tasks.} This task requires leveraging dynamic tactile cues for reasoning. TacReasoner achieves consistent improvements over VTV-LLM on both RFOR and OCSE tasks.}
\label{tab2}
\resizebox{0.85\columnwidth}{!}{
\begin{tabular}{lccc}
\hline
 & Random & VTV-LLM-7B & \textbf{TacReasoner-7B} \\
\hline
RFOR & 50.00 & 43\% & \textbf{68\%} \\
OCSE & 33.33 & 46\% & \textbf{53\%} \\
\hline
\end{tabular}
}
\vspace{-2mm}
\end{table}

\begin{figure}[!t] 
    \centering 
    \includegraphics[width=\columnwidth]{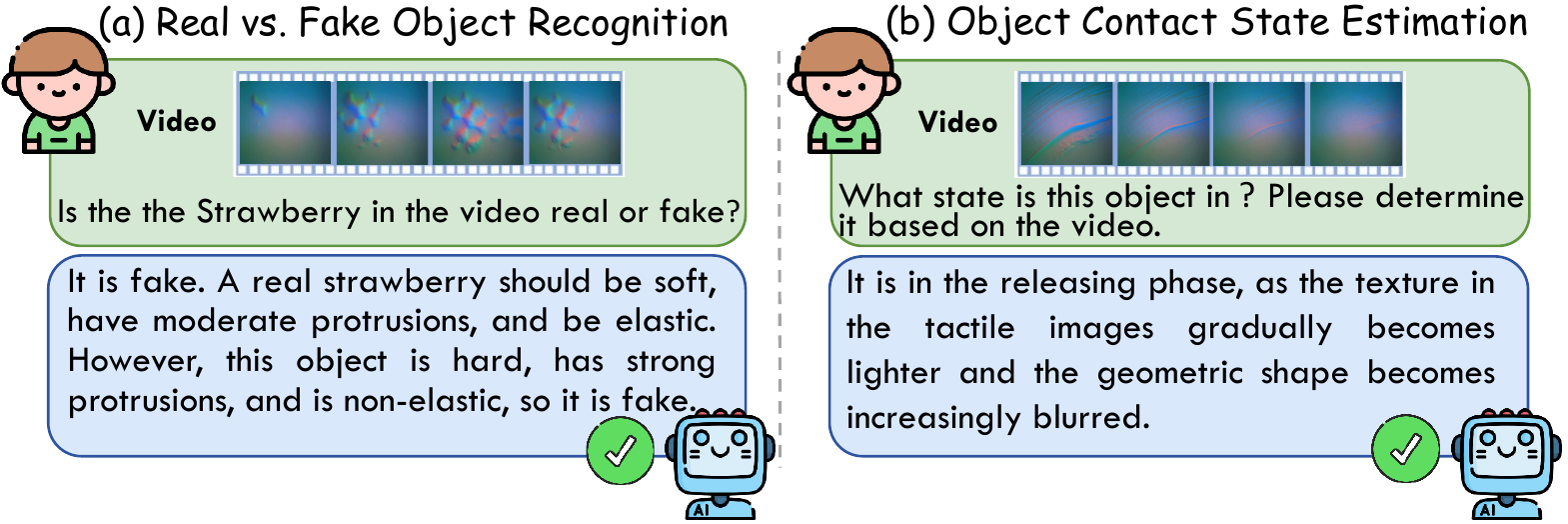} 
    \caption{\textbf{The reasoning examples of TacReasoner on tactile dynamic-aware reasoning tasks.} TacReasoner achieves accurate reasoning across all tasks.}
    \vspace{-4mm}
    \label{exampe2}
\end{figure}

\subsection{Tactile Dynamic-Aware Reasoning}
\label{Tactile Dynamic-Aware Reasoning}
As stated in Section~\ref{intro}, the essence of tactile perception lies in capturing dynamic information and performing reasoning. We evaluate this capability using the Real vs. Fake Object Recognition \textit{(RFOR)} and Object Contact State Estimation \textit{(OCSE)} tasks in DynTAC-Bench. For each task, 3 samples are randomly selected per category. For Object Contact State Estimation, state transition intervals (e.g., contact, sliding, and release) are randomly segmented for evaluation, yielding 45 test samples in total. These samples are evaluated using TacReasoner-7B and VTV-LLM-7B \cite{vtv}, with results reported in Table~\ref{tab2}. TacReasoner improves accuracy over VTV-LLM-7B by 25\% on RFOR task and 7\% on OCSE task, respectively, and substantially outperforms the random baseline. These results demonstrate that TacReasoner effectively captures dynamic tactile variations and performs reasoning. Figures~\ref{exampe2}(a) and (b) further provide qualitative examples of TacReasoner in tactile dynamic aware reasoning.

\begin{table}[!t]
\centering
\renewcommand{\arraystretch}{1.1}
\setlength{\tabcolsep}{3pt}
\caption{Ablation study on DynEncoder and TouchCoT-10K SFT settings using the TacReasoner-7B model.}
\label{tab3}
\resizebox{\columnwidth}{!}{
\begin{tabular}{cc|ccccc}
\hline
\multicolumn{2}{c|}{Setting} & \multicolumn{5}{c}{Accuracy (\%)} \\
\cline{1-2} \cline{3-7}
DynEncoder & 
\begin{tabular}{c}
SFT \\
w/ TouchCoT-10K
\end{tabular}
& SFD & SOI & OSC & TSA & Average \\
\hline
 &  & 71.3 & 57.6 & 43.2 & 64.0 & 59.0 \\
  $\checkmark$&  & 72.4 & 64.5 & 48.6 & 66.0 & 62.9 \\
 & $\checkmark$ & 75.2  & 68.5 & 50.7 & 69.0 & 65.8 \\
 $\checkmark$ & $\checkmark$ 
 & \textbf{73.6} & \textbf{70.3} & \textbf{54.3} & \textbf{71.0} & \textbf{67.3} \\
\hline
\end{tabular}
}
\vspace{-1mm}
\end{table}

\begin{table}[!t]
\centering
\renewcommand{\arraystretch}{1.1}
\caption{Ablation study on training paradigm settings using the TacReasoner-7B model.}
\label{tab4}
\resizebox{0.95\columnwidth}{!}{
\begin{tabular}{l|cccc|c}
\hline
Settings & SFD & SOI & OSC & TSA & Average \\
\hline
w/o stage 1 & 56.8 & 58.4 & 32.6 & 56.0 & 51.0 \\
w/o stage 2 & 52.7 & 43.8 & 27.6 & 39.6  & 40.9 \\
VTV-150K & 72.4 & 64.5 & 48.6 & 66.0 & 62.9 \\
\textbf{Ours} & \textbf{73.6} & \textbf{70.3} & \textbf{54.3} & \textbf{71.0} & \textbf{67.3} \\
\hline
\end{tabular}
}
\label{trainpar}
\vspace{-4mm}
\end{table}

\subsection{Ablation Studies}
\label{Ablation Studies}
In this section, we conduct ablation studies to analyze the impact of key factors in the TacReasoner model, including model scale, the dynamic-aware tactile encoder, TouchCoT 10K, and the training strategy, as detailed below.

\noindent \textbf{Impact of the model scale.} To investigate the impact of model scale on visuo-tactile understanding, we evaluate LLM backbones of varying sizes and report the performance of TacReasoner with Qwen-2.5-14B \cite{qwen}, as shown in Table~\ref{tab1}. The results demonstrate consistent improvements across most subtasks as model size increases, with especially significant improvements on reasoning-oriented tasks such as SFD.

\noindent \textbf{Impact of the \underline{Dyn}amic-aware Tactile \underline{Encoder} (DynEncoder).} We perform an ablation study on DynEncoder, as reported in Table~\ref{tab3}. VTV-LLM models tactile dynamics via video reconstruction but operates on partial regions without conditional guidance, limiting dynamic representation. DynEncoder introduces inter-frame dynamic modeling and question-guided conditioning, enabling query-relevant tactile reasoning. It achieves consistent gains on reasoning tasks and improves average performance by 4.6\%, validating its effectiveness in enhancing tactile understanding.

\noindent \textbf{Impact of the Training Paradigm and TouchCoT 10K.} Table~\ref{trainpar} validates our two-stage training paradigm via ablation study. Removing Stage 1 misaligns tactile and text tokens, reducing average performance to 51\%. Omitting Stage 2 removes end-to-end fine-tuning, causing unstable LLM token generation and a sharp performance drop to 40.9\%. These results confirm that progressive training is essential for improving model abilities. As shown in Tables~\ref{tab3} and ~\ref{tab4}, training with TouchCoT-10K and VTV-150K under the same paradigm shows that TouchCoT-10K brings significant gains, validating its efficacy. This benefit stems from CoT data supervising and activating the LLM’s reasoning process.

\section{Conclusion and Future work}
\label{Conclusion and Future work}
In this work, we present \textbf{TacReasoner}, a dynamic tactile-language reasoning framework for real-world scenarios that mitigates inadequate modeling of contact temporal evolution and semantic hallucination. TacReasoner explicitly captures temporal dynamics and causal dependencies during interaction through the Dynamic-aware Tactile Encoder, producing enhanced tactile embeddings tailored for reasoning. To support structured reasoning, we construct the first chain-of-thought dataset for tactile reasoning, \textbf{TouchCoT-10K}, and employ supervised fine-tuning to align dynamic tactile evidence with structured attribute-level conclusions. Furthermore, We introduce \textbf{DynTAC-Bench} to evaluate performance across diverse tasks, especially dynamic-aware reasoning. Extensive experiments show that TacReasoner outperforms strong baselines across datasets and reasoning tasks, demonstrating its effectiveness for dynamic tactile understanding and decision making. Future work will further improve interaction diversity, cross-sensor and cross-scenario generalization, and deployment on real robotic systems.

\bibliographystyle{IEEEtran} 
\bibliography{references}  

@article{tactile1,
  title={Why is there so much more research on vision than on any other sensory modality?},
  author={Hutmacher, Fabian},
  journal={Frontiers in psychology},
  volume={10},
  pages={481030},
  year={2019},
  publisher={Frontiers}
}

@article{tactile2,
  title={The sense of touch},
  author={O'Shaughnessy, Brian},
  journal={Australasian journal of philosophy},
  volume={67},
  number={1},
  pages={37--58},
  year={1989},
  publisher={Taylor \& Francis}
}

@article{human11,
  title={Human tactile perception as a standard for artificial tactile sensing—a review},
  author={Dargahi, Javad and Najarian, Siamak},
  journal={The international journal of medical robotics and computer assisted surgery},
  volume={1},
  number={1},
  pages={23--35},
  year={2004},
  publisher={Wiley Online Library}
}

@article{tacsensor2,
  title={Tactile sensors: A review},
  author={Meribout, Mahmoud and Takele, Natnael Abule and Derege, Olyad and Rifiki, Nidal and El Khalil, Mohamed and Tiwari, Varun and Zhong, Jing},
  journal={Measurement},
  volume={238},
  pages={115332},
  year={2024},
  publisher={Elsevier}
}

@article{tacsensor3,
  title={Recent progress in tactile sensors and their applications in intelligent systems},
  author={Liu, Yue and Bao, Rongrong and Tao, Juan and Li, Jing and Dong, Ming and Pan, Caofeng},
  journal={Science Bulletin},
  volume={65},
  number={1},
  pages={70--88},
  year={2020},
  publisher={Elsevier}
}

@inproceedings{touchformer,
  title={TouchFormer: A Robust Transformer-based Framework for Multimodal Material Perception},
  author={Lyu, Kailin and Xiao, Long and Zeng, Jianing and Dong, Junhao and Liu, Xuexin and Zou, Zhuojun and Yang, Haoyue and Shu, Lin and Hao, Jie},
  booktitle={Proceedings of the AAAI Conference on Artificial Intelligence},
  volume={40},
  number={22},
  pages={18496--18504},
  year={2026}
}

@article{taob,
  title={Touch and go: Learning from human-collected vision and touch},
  author={Yang, Fengyu and Ma, Chenyang and Zhang, Jiacheng and Zhu, Jing and Yuan, Wenzhen and Owens, Andrew},
  journal={arXiv preprint arXiv:2211.12498},
  year={2022}
}

@inproceedings{clip,
  title={Learning transferable visual models from natural language supervision},
  author={Radford, Alec and Kim, Jong Wook and Hallacy, Chris and Ramesh, Aditya and Goh, Gabriel and Agarwal, Sandhini and Sastry, Girish and Askell, Amanda and Mishkin, Pamela and Clark, Jack and others},
  booktitle={International conference on machine learning},
  pages={8748--8763},
  year={2021},
  organization={PmLR}
}

@article{tacman2,
  title={A survey of robot manipulation in contact},
  author={Suomalainen, Markku and Karayiannidis, Yiannis and Kyrki, Ville},
  journal={Robotics and Autonomous Systems},
  volume={156},
  pages={104224},
  year={2022},
  publisher={Elsevier}
}

@article{mllm,
  title={Surveying the mllm landscape: A meta-review of current surveys},
  author={Li, Ming and Chen, Keyu and Bi, Ziqian and Liu, Ming and Song, Xinyuan and Jiang, Zekun and Wang, Tianyang and Peng, Benji and Niu, Qian and Liu, Junyu and others},
  journal={arXiv preprint arXiv:2409.18991},
  year={2024}
}

@article{octopi,
  title={Octopi: Object property reasoning with large tactile-language models},
  author={Yu, Samson and Lin, Kelvin and Xiao, Anxing and Duan, Jiafei and Soh, Harold},
  journal={arXiv preprint arXiv:2405.02794},
  year={2024}
}

@article{octopi15,
  title={Demonstrating the octopi-1.5 visual-tactile-language model},
  author={Yu, Samson and Lin, Kelvin and Soh, Harold},
  journal={arXiv preprint arXiv:2507.09985},
  year={2025}
}

@article{touchthinker,
  title={TouchThinker: Scaling Tactile Commonsense Reasoning to the Open World with Large-scale Data and Action-aware Representation},
  author={Lyu, Kailin and Wu, Di and Zhang, Pengwei and Zheng, Yuhang and Lai, Yingxin and Xiao, Long and Wu, Kangyi and Li, Pengna and Gao, Chen and Hu, Lianyu and others},
  journal={arXiv preprint arXiv:2606.11637},
  year={2026}
}

@article{th2,
  title={Restoring tactile and proprioceptive sensation through a brain interface},
  author={Tabot, Gregg A and Kim, Sung Shin and Winberry, Jeremy E and Bensmaia, Sliman J},
  journal={Neurobiology of disease},
  volume={83},
  pages={191--198},
  year={2015},
  publisher={Elsevier}
}

@inproceedings{unitouch,
  title={Binding touch to everything: Learning unified multimodal tactile representations},
  author={Yang, Fengyu and Feng, Chao and Chen, Ziyang and Park, Hyoungseob and Wang, Daniel and Dou, Yiming and Zeng, Ziyao and Chen, Xien and Gangopadhyay, Rit and Owens, Andrew and others},
  booktitle={Proceedings of the IEEE/CVF Conference on Computer Vision and Pattern Recognition},
  pages={26340--26353},
  year={2024}
}

@inproceedings{tacgrasp,
  title={Learning of grasp adaptation through experience and tactile sensing},
  author={Li, Miao and Bekiroglu, Yasemin and Kragic, Danica and Billard, Aude},
  booktitle={2014 IEEE/RSJ International Conference on Intelligent Robots and Systems},
  pages={3339--3346},
  year={2014},
  organization={Ieee}
}

@inproceedings{grasp2,
  title={Robotic grasping and contact: A review},
  author={Bicchi, Antonio and Kumar, Vijay},
  booktitle={Proceedings 2000 ICRA. Millennium conference. IEEE international conference on robotics and automation. Symposia proceedings (Cat. No. 00CH37065)},
  volume={1},
  pages={348--353},
  year={2000},
  organization={IEEE}
}

@inproceedings{tacin,
  title={Tactile-based insertion for dense box-packing},
  author={Dong, Siyuan and Rodriguez, Alberto},
  booktitle={2019 IEEE/RSJ International Conference on Intelligent Robots and Systems (IROS)},
  pages={7953--7960},
  year={2019},
  organization={IEEE}
}

@article{t3,
  title={Transferable tactile transformers for representation learning across diverse sensors and tasks},
  author={Zhao, Jialiang and Ma, Yuxiang and Wang, Lirui and Adelson, Edward H},
  journal={arXiv preprint arXiv:2406.13640},
  year={2024}
}

@article{anytouch,
  title={Anytouch: Learning unified static-dynamic representation across multiple visuo-tactile sensors},
  author={Feng, Ruoxuan and Hu, Jiangyu and Xia, Wenke and Gao, Tianci and Shen, Ao and Sun, Yuhao and Fang, Bin and Hu, Di},
  journal={arXiv preprint arXiv:2502.12191},
  year={2025}
}

@article{deepseek,
  title={Exploring DeepSeek: A survey on advances, applications, challenges and future directions},
  author={Deng, Zehang and Ma, Wanlun and Han, Qing-Long and Zhou, Wei and Zhu, Xiaogang and Wen, Sheng and Xiang, Yang},
  journal={IEEE/CAA Journal of Automatica Sinica},
  volume={12},
  number={5},
  pages={872--893},
  year={2025},
  publisher={IEEE}
}

@article{lora,
  title={When scaling meets llm finetuning: The effect of data, model and finetuning method},
  author={Zhang, Biao and Liu, Zhongtao and Cherry, Colin and Firat, Orhan},
  journal={arXiv preprint arXiv:2402.17193},
  year={2024}
}

@article{vtv,
  title={Universal visuo-tactile video understanding for embodied interaction},
  author={Xie, Yifan and Li, Mingyang and Li, Shoujie and Li, Xingting and Chen, Guangyu and Ma, Fei and Yu, Fei Richard and Ding, Wenbo},
  journal={arXiv preprint arXiv:2505.22566},
  year={2025}
}

@article{adamw,
  title={Towards understanding convergence and generalization of AdamW},
  author={Zhou, Pan and Xie, Xingyu and Lin, Zhouchen and Yan, Shuicheng},
  journal={IEEE transactions on pattern analysis and machine intelligence},
  volume={46},
  number={9},
  pages={6486--6493},
  year={2024},
  publisher={IEEE}
}

@article{qwen,
  title={Qwen-image technical report},
  author={Wu, Chenfei and Li, Jiahao and Zhou, Jingren and Lin, Junyang and Gao, Kaiyuan and Yan, Kun and Yin, Sheng-ming and Bai, Shuai and Xu, Xiao and Chen, Yilei and others},
  journal={arXiv preprint arXiv:2508.02324},
  year={2025}
}

@article{gpt4o,
  title={Gpt-4o system card},
  author={Hurst, Aaron and Lerer, Adam and Goucher, Adam P and Perelman, Adam and Ramesh, Aditya and Clark, Aidan and Ostrow, AJ and Welihinda, Akila and Hayes, Alan and Radford, Alec and others},
  journal={arXiv preprint arXiv:2410.21276},
  year={2024}
}

@article{gemini,
  title={Gemini 2.5: Pushing the frontier with advanced reasoning, multimodality, long context, and next generation agentic capabilities},
  author={Comanici, Gheorghe and Bieber, Eric and Schaekermann, Mike and Pasupat, Ice and Sachdeva, Noveen and Dhillon, Inderjit and Blistein, Marcel and Ram, Ori and Zhang, Dan and Rosen, Evan and others},
  journal={arXiv preprint arXiv:2507.06261},
  year={2025}
}

@article{llava,
  title={Llava-onevision: Easy visual task transfer},
  author={Li, Bo and Zhang, Yuanhan and Guo, Dong and Zhang, Renrui and Li, Feng and Zhang, Hao and Zhang, Kaichen and Zhang, Peiyuan and Li, Yanwei and Liu, Ziwei and others},
  journal={arXiv preprint arXiv:2408.03326},
  year={2024}
}

@article{llava-video,
  title={Video instruction tuning with synthetic data, 2024},
  author={Zhang, Yuanhan and Wu, Jinming and Li, Wei and Li, Bo and Ma, Zejun and Liu, Ziwei and Li, Chunyuan},
  journal={URL https://arxiv. org/abs/2410.02713},
  volume={17}
}

@article{internvl,
  title={Expanding performance boundaries of open-source multimodal models with model, data, and test-time scaling},
  author={Chen, Zhe and Wang, Weiyun and Cao, Yue and Liu, Yangzhou and Gao, Zhangwei and Cui, Erfei and Zhu, Jinguo and Ye, Shenglong and Tian, Hao and Liu, Zhaoyang and others},
  journal={arXiv preprint arXiv:2412.05271},
  year={2024}
}

@article{videollama,
  title={Videollama 3: Frontier multimodal foundation models for image and video understanding},
  author={Zhang, Boqiang and Li, Kehan and Cheng, Zesen and Hu, Zhiqiang and Yuan, Yuqian and Chen, Guanzheng and Leng, Sicong and Jiang, Yuming and Zhang, Hang and Li, Xin and others},
  journal={arXiv preprint arXiv:2501.13106},
  year={2025}
}

\end{document}